\documentclass[runningheads]{llncs}
\usepackage{graphicx}
\usepackage{amsmath,amssymb} 
\usepackage{color}
\usepackage[width=122mm,left=12mm,paperwidth=146mm,height=193mm,top=12mm,paperheight=217mm]{geometry}

\usepackage{subfig}

\usepackage{epsfig}
\usepackage{amsmath}
\usepackage{amssymb}
\usepackage{multirow}
\usepackage{color}
\usepackage{array}
\newcolumntype{L}[1]{>{\raggedright\let\newline\\\arraybackslash\hspace{0pt}}m{#1}}
\newcolumntype{C}[1]{>{\centering\let\newline\\\arraybackslash\hspace{0pt}}m{#1}}
\newcolumntype{R}[1]{>{\raggedleft\let\newline\\\arraybackslash\hspace{0pt}}m{#1}}
\usepackage{mathtools}

\newcommand*{\affmark}[1][*]{\textsuperscript{#1}}

\begin{document}
\pagestyle{headings}
\mainmatter
\def\ECCV18SubNumber{50}  

\title{Factorized Adversarial Networks for Unsupervised Domain Adaptation} 

\titlerunning{Factorized Adversarial Networks for Unsupervised Domain Adaptation}

\authorrunning{Jian Ren \textit{et al.}}

\author{Jian Ren\affmark[1], Jianchao Yang\affmark[2], Ning Xu\affmark[3], David J. Foran\affmark[1]}

\institute{\affmark[1]Rutgers University, \affmark[2]Toutiao AI Lab,  \affmark[3]Snap Research}

\maketitle

\begin{abstract}
In this paper, we propose Factorized Adversarial Networks (FAN) to solve unsupervised domain adaptation problems for image classification tasks. Our networks map the data distribution into a latent feature space, which is factorized into a domain-specific subspace that contains domain-specific characteristics and a task-specific subspace that retains category information, for both source and target domains, respectively. Unsupervised domain adaptation is achieved by adversarial training to minimize the discrepancy between the distributions of two task-specific subspaces from source and target domains.
We demonstrate that the proposed approach outperforms state-of-the-art methods on multiple benchmark datasets used in the literature for unsupervised domain adaptation. Furthermore, we collect two real-world tagging datasets that are much larger than existing benchmark datasets, and get significant improvement upon baselines, proving the practical value of our approach.
\end{abstract}
\section{Introduction}

Rapid development of deep convolutional neural networks (CNN) has led to promising performance on various computer vision tasks~\cite{krizhevsky2012imagenet}\cite{szegedy2015going}\cite{szegedy2013deep}, especially with the help of large-scale annotated datasets, such as ImageNet~\cite{deng2009imagenet}. However, when a model learned from a large dataset in one domain (source domain) is applied to another domain (target domain) with some different characteristics, it is not guaranteed to generalize well. In order to mitigate the influence caused by domain shift~\cite{gretton2009covariate}, two major approaches are widely employed. One popular approach is to fine-tune the model learned from source domain using annotated data from target distribution~\cite{schmidhuber2015deep}. However, this requires data annotation in target domain, which is costly and labor intensive. The other approach is to generate synthetic data that is analogous to the distribution of target domain~\cite{bousmalis2016unsupervised}\cite{shrivastava2016learning}. Although this approach could provide unlimited synthetic training data, the model trained may not perform well as compared to real data with much more complicated distributions. 

\begin{figure}
\begin{center}
\includegraphics[width=0.6\columnwidth]{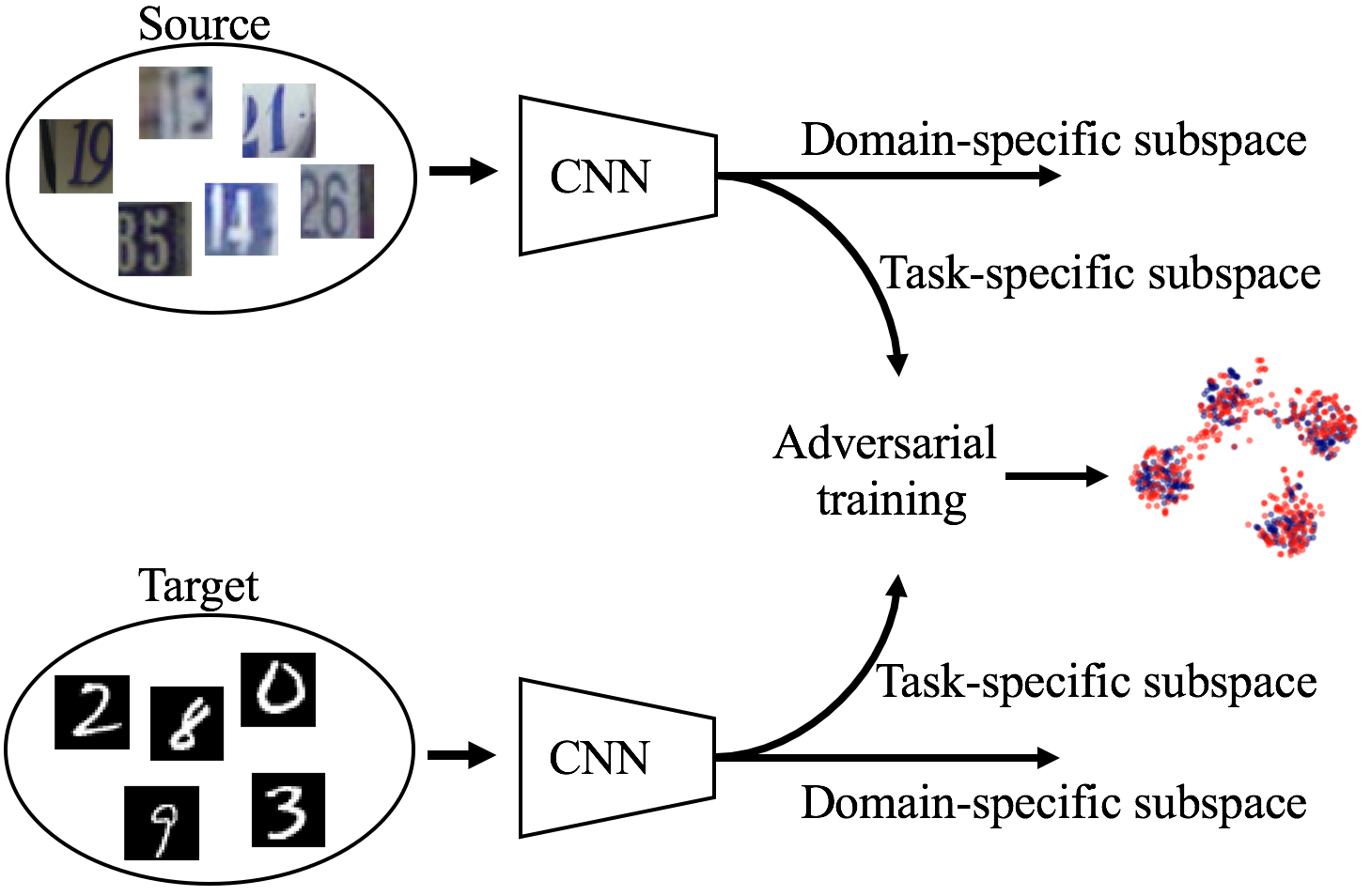}
\caption{The proposed unsupervised domain adaptation approach factorizes source and target latent feature space into two subspaces using two different networks. The domain-specific subspace stores domain-specific information, while the task-specific subspace stores the category information. We use adversarial training to minimize the discrepancy between the two task-specific subspaces.}
\label{outline_figure}
\end{center}
\vspace*{-7mm}
\end{figure}

In this work, we focus on the image classification task and aim to solve the unsupervised domain adaptation problem. In our problem setting, the source domain contains a large amount of annotated data, but there is no annotation available for the images in the target domain. The two domains share the same high level categories although they are drawn from different distributions. 


We propose Factorized Adversarial Networks (FAN) to address this unsupervised domain adaptation problem. FAN encodes input data from both domains to a latent embedding space which is factorized into two complementary subspaces, a domain-specific subspace (DSS) and a task-specific subspace (TSS), as illustrated in Figure~\ref{outline_figure}. In an image recognition scenario, the task-specific subspace should ideally only contain image category related information, while the domain-specific subspace contains domain characteristics that are irrelevant to classification, e.g., different backgrounds should not impact digit recognition. We use a mutual information loss to enforce the orthogonality constraint between the two subspaces. The motivation of this factorization is to allow us to adapt only the task-specific subspace of the target domain to that of the source domain. In order to do the adaptation, we apply an adversarial network to minimize the distribution discrepancy between the two task-specific subspaces, with loss function adopted from the Generative Adversarial Network (GAN)~\cite{goodfellow2014generative}. 

A two-stage training process is used to train our FAN. In the first stage, we train a convolutional network in source domain to predict the image labels as well as reconstruct the input images. The features in task-specific subspace are used to predict the image labels, while the domain-specific subspace features, concatenated with the image classification logits, are used to reconstruct the input images. In the second stage, we train the network in target domain using the adversarial loss and reconstruction loss to generate a task-specific subspace that is indistinguishable from the one generated in source domain. A discriminator network is used to judge from which domain the task-specific features are generated. The network at target domain and the discriminator network are updated by the gradients in an adversarial way so that the task-specific subspace of target domain is adapted to that subspace of source domain. 

We apply our proposed method to visual domain adaptation using the benchmark digits datasets, including MNIST~\cite{lecun1998gradient}, USPS~\cite{hull1994database} and SVHN~\cite{netzer2011reading}, and achieve superior results compared to the state-of-the-art approaches. We also apply the method to two real-world tagging datasets that we collected, one from crawling images using search engines such as Google and Flickr, and the other from photos shot by mobile phones. The two datasets share the same 100 classes with each dataset containing more than 115,000 images and we achieve significant improvement on the classification task compared with the state-of-the-arts. 

In summary, our contributions are three-fold:
\begin{itemize}
\item A novel Factorized Adversarial Networks to tackle the unsupervised domain adaptation in an effective way.
\item Detailed analysis on the design of the network architecture along with visualization of the factorized subspaces.
\item New state-of-the-art domain adaptation results on digits benchmark datasets as well as newly collected larger-scale real-world tagging datasets.
\end{itemize}

\section{Related Work}

{\textbf{Unsupervised domain adaptation}} Extensive studies on unsupervised domain adaptation have been conducted in recent years in order to effectively transfer the representative features learned in source domain to target domain. In this section, we focus on research utilizing deep neural networks as they have a better generalization ability even for the complex distributions~\cite{krizhevsky2012imagenet}\cite{He2015}\cite{luo2017label}. 

One category of unsupervised domain adaptation applies the Maximum Mean Discrepancy (MMD)~\cite{gretton2012kernel} loss as a metric to learn the domain invariant features. The MMD loss computes the distance between the embedding spaces of two domains using kernel tricks. Deep domain confusion (DDC)~\cite{tzeng2014deep} minimizes both classification loss and MMD loss in one layer. Deep adaptation network proposed in~\cite{long2015learning} places MMD loss at multiple task-specific layers that have been embedded in a reproducing kernel Hilbert space, while other layers are shared between source and target domains. Similarly, the domain separation network (DSN)~\cite{bousmalis2016domain} maintains a shared embedding between two domains as well as the individual domain representations. Deep Reconstruction-Classification Network (DRCN)~\cite{ghifary2016deep} shares the encoding for both source and target domains. On the contrary, the work in~\cite{rozantsev2016beyond} demonstrates that it is effective to relate the weights in the form of linear transformations instead of sharing. Unlike the above discussed approaches, the authors in~\cite{sun2016deep} proposed deep correlation alignment (CORAL) algorithm to match the covariance of the source features and the target features to learn a transformation from the source domain to the target domain. 

Based on the idea of adversarial training~\cite{goodfellow2014generative}, several studies propose using a domain classifier built on top of the networks to distinguish the represented features from the two distributions. Features extracted from the two domains are utilized to train the domain classifier, along with the classification loss for the source domain~\cite{tzeng2015simultaneous}. The gradient reversal algorithm (RevGrad) algorithm~\cite{ganin2015unsupervised} trains the domain classifier by reversing its gradients. The authors of~\cite{tzeng2017adversarial} propose an adversarial discriminative domain adaptation (ADDA) model in which weights are not shared between the source and target domains, and the network in target domain is trained to fool the domain classifier so that it cannot predict the two domains reliably.

{\textbf{Generative adversarial networks}}
GAN~\cite{goodfellow2014generative} related approaches are also used to synthesize images and perform unsupervised domain adaptation in the joint distribution space. A generator is trained to model the image distribution and generate the synthetic images while a discriminator is trained to differentiate the synthesized distribution and the real distribution. Coupled GAN (CoGAN)~\cite{liu2016coupled} uses two GANs on source and target domain to generate images from the two distributions. The two GANs have the same noise as input and domain adaptation is implemented by training a classifier on the input of the discriminator. The work in~\cite{bousmalis2016unsupervised} uses images from source domain as a condition for the generator. Both the generated images and the source images are applied to train the classifier. The authors of~\cite{taigman2016unsupervised} propose a learning strategy to generate cross domain images and train a task-specific classifier with the generated images and the source distributions.

{\textbf{Hidden factors discovery}}
There has been some research work on discovering the higher-order factors of variation from the latent space on the image classification and generation tasks~\cite{cheung2014discovering}\cite{chen2016infogan}\cite{mathieu2016disentangling}\cite{makhzani2015adversarial}. For example, the work at~\cite{cheung2014discovering} utilizes the autoencoder to disentangle the various transformations from input distributions. The network is jointly trained to reconstruct input images as well as estimate the image category. On the contrary, InfoGAN~\cite{chen2016infogan} is proposed to learn disentangled representations from images in an unsupervised fashion by decomposing the latent code from input noise vector. In this study, we propose learning the task-specific feature in an effective way instead of learning interpretable hidden factors, and we find that factorizing the domain representations helps to adapt the knowledge between two domains.

{\textbf{Comparison with similar studies}}
The motivation of our proposed FAN is to find a subspace where unsupervised domain adaptation for classification is most appropriate. It shares similarities with previous studies, especially DSN~\cite{bousmalis2016domain} and the ADDA~\cite{tzeng2017adversarial}. While domain separation~\cite{cheung2014discovering}\cite{salzmann2010factorized}\cite{jia2010factorized} and adversarial training~\cite{bousmalis2016unsupervised}\cite{tzeng2015simultaneous} have been extensively explored in many tasks in existing liturature, we unify the two appoaches in one novel framework for unsupervised domain adaptation and demonstrate its clear advantage over DSN~\cite{bousmalis2016domain} and ADDA~\cite{tzeng2017adversarial} in experiments.  


\begin{figure*}[t!]
\begin{center}
\includegraphics[width=1\columnwidth]{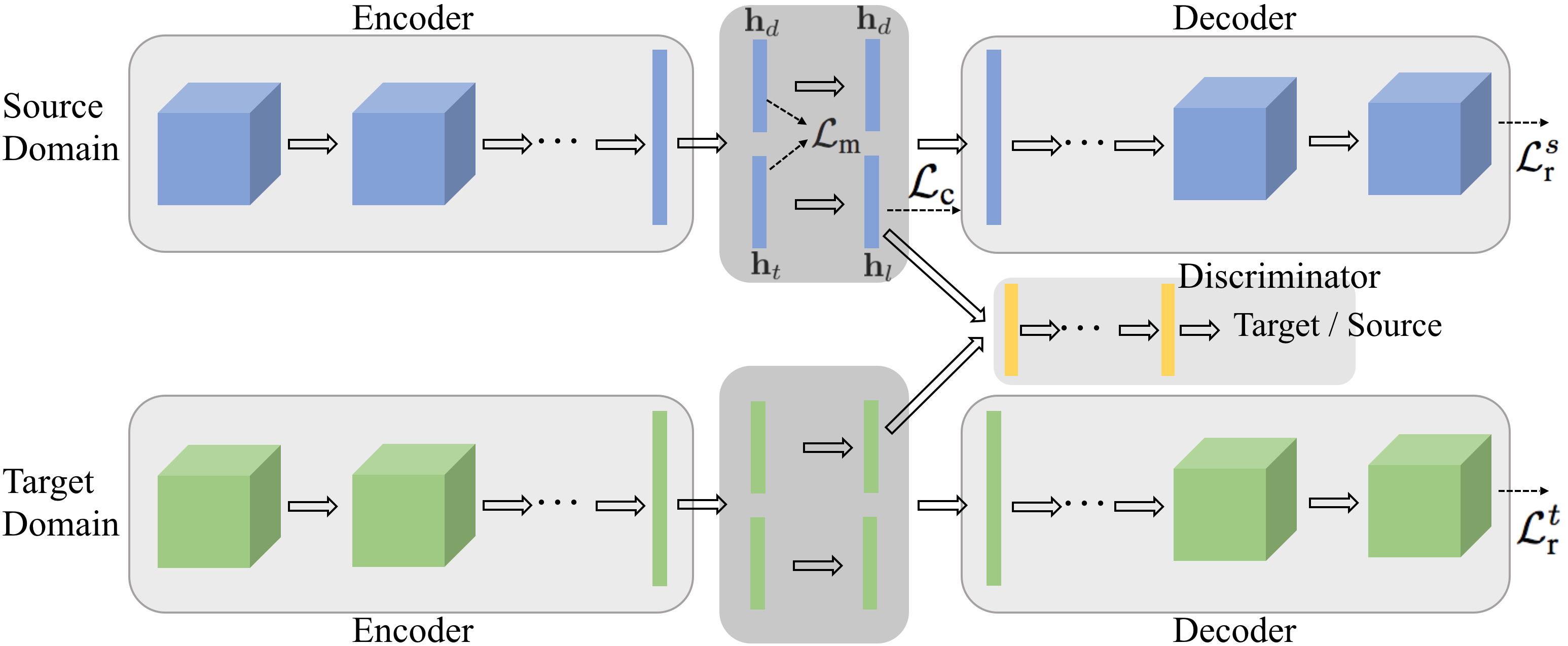}
\caption{The architecture of FAN. 
The encoders from two domains map input images into two feature spaces. Both feature spaces are factorized into two subspaces, the domain-specific subspace (DSS) and the task-specific subspace (TSS). The adaptation is accomplished by jointly training the discriminator and target network using both the GAN loss and reconstruction loss to find the domain invariant feature in TSS.}
\label{network}
\end{center}
\vspace*{-5mm}
\end{figure*}

\section{Our Approach}

In this section, we present our Factorized Adversarial Networks (FAN) for unsupervised domain adaptation. The architecture of FAN is illustrated in Figure~\ref{network}, where we have two encoder-decoder structured neural networks, one for source domain and one for target domain, that mirror each other except for the training losses, as well as a discriminator network. We aim to find a domain invariant feature space that retains the classification information through adversarial training. To achieve this, we explicitly factor the latent feature space into a task-specific subspace and a complementary domain-specific subspace, where the task-specific subspace aims to minimize the classification loss across domains while the domain-specific subspace combined with classification logits targets at reconstructing the input samples. The task-specific subspace, if indistinguishable by the adversarial discriminator which domain it comes from, should retain the classification information invariant to domain shifts; the domain-specific subspace, on the other hand, should capture the domain-specific but classification-irrelevant information for reconstruction. The proposed explicit feature space factorization helps to remove some domain-specific information and relieve the burden of adversarial training for more effective domain adaptation.

More formally, in our unsupervised domain adaptation, we have a source distribution $\mathcal{S}$ that includes $N^s$ labeled images $\left \{ {(\mathbf{x}_{i}^s, \mathbf{y}_{i}^s)} \right \}_{i=1}^{N^s}$ where $\mathbf{y}_{i}^s$ is a one-hot vector encoding the image class label, and a target distribution $\mathcal{T}$ contains $N^t$ unlabeled images $\left \{ {(\mathbf{x}_{i}^t)} \right \}_{i=1}^{N^t}$. Our goal is to first find a mapping $M^s$ that maps the source task-specific subspace to the source logit space with labeled training data, and then find a mapping function $M^t$ for the target domain that maps the target task-specific subspace into the target logit space that is indistinguishable from the source logit space. The target mapping function $M^t$ thus retains the discriminative information needed for target domain, and therefore, inference in target domain could be easily done with $M^t$ and~\textit{softmax}. Our learning procedure consists of two steps: we first train a source domain network that factors the latent feature space, and we then update the target domain network by adapting the target domain task-specific subspace to its source domain counterpart with the help of adversarial training. We discuss these two steps in the following sections.

\subsection{Feature Space Factorization}
Our networks contain two convolutional encoder-decoder networks and the latent feature space generated by the encoders is factorized into
complementary task-specific subspace and domain-specific subspace. In the first step of our approach, we train our factorization network in source domain as shown in Figure~\ref{network}. To avoid cluttered notations, we drop domain indicator superscripts $s$ in the following when there is no confusion. Let $\mathbf{h}$ = $Enc(\mathbf{x}; \theta_e)$ denote the encoder function $Enc$ that encodes the input sample $\mathbf{x}$ into a latent feature $\mathbf{h}$ with parameter $\theta_e$ in source domain. We split the latent feature $\mathbf{h}$ into two parts $\mathbf{h}_d$ and $\mathbf{h}_t$, where $\mathbf{h}_d$ represents the feature in the domain-specific subspace and $\mathbf{h}_t$ represents the feature in the task-specific subspace. The mapping $\mathbf{h}_l = M(\mathbf{h}_t; \theta_m)$ maps the task-specific subspace into a logit space with parameters $\theta_m$. We then concatenate $\mathbf{h}_d$ and $\mathbf{h}_l$ and feed them into a decoder $Dec(\mathbf{h}_d, \mathbf{h}_l; \theta_d)$ to reconstruct the input sample $\mathbf{x}$, where $\mathbf{h}_l$ includes the necessary attributes for reconstruction. Ideally, $\mathbf{h}_t$ should contain discriminant information that is invariant to different domains while $\mathbf{h}_d$ retains information that is specific to the domain, less relevant to classification but necessary for reconstruction. We optimize the following objective function in order to obtain the two desired subspaces in source domain:
\begin{equation}
\mathcal{L}_\text{source} = \alpha \mathcal{L}_\text{c} + \beta \mathcal{L}_\text{m} + \mathcal{L}_{\text{r}}
\label{loss_total}
\end{equation}
where $\alpha$, $\beta$ are hyper parameters that control the trade-off among loss terms. 

$\mathcal{L}_\text{c}$ is the cross-entropy loss to train the source network for classification with the parameters $\left \{ \theta_e, \theta_m \right \}$ using source domain labeled training data. 
\begin{equation}
\mathcal{L}_\text{c} = -\sum_{i=1}^{N}\mathbf{y}_i \cdot \text{log} \hspace{0.1cm}\hat{\mathbf{y}}_i
\label{task_loss}
\end{equation}
where $\hat{\mathbf{y}}$ is the \textit{softmax} output of the classification branch, $\hat{\mathbf{y}}$ = \textit{softmax}$(M(\mathbf{h}_t; \theta_m))$. 

We add a mutual information loss term $\mathcal{L}_\text{m}$ to encourage orthogonality between the domain-specific subspace and task-specific subspace: 
\begin{equation}
\mathcal{L}_\text{m} = \sum_{i=1}^{N} \left \| \mathbf{h}_{ti}^{\textbf{T}} \mathbf{h}_{di} \right \|^{2}
\label{mutal_loss}
\end{equation}
where $\mathbf{h}_{ti}$ and $\mathbf{h}_{di}$ denote the domain-specific feature and task-specific feature for the $i$-th sample, respectively.

We use the reconstruction loss $\mathcal{L}_{\text{r}}$ to minimizes the squared error between the input sample and the reconstructed one:
\begin{equation}
 \mathcal{L}_{\text{r}} = \sum_{i=1}^{N} \left \| \mathbf{x}_i - Dec(\mathbf{h}_{di}, \mathbf{h}_{li}; \theta_d) \right \|^{2}
 \label{recon_loss}
\end{equation}
where $\mathbf{h}_{li}$ denote the  logit vector for the $i$-th sample.

The three loss terms play together in the optimization of Eqn.~\ref{loss_total}. The classification loss $\mathcal{L}_\text{c}$ encourages the learned feature $\mathbf{h}_t$ to retain discriminative information as much as possible, the reconstruction loss $\mathcal{L}_\text{r}$ relies on domain-specific information from $\mathbf{h}_d$ with the logit input $\mathbf{h}_l$ for reconstruction, and the mutual loss $\mathcal{L}_\text{m}$ encourages the separation of the two subspaces. Thus we can obtain a task-specific space $\mathbf{h}_t$ that is discriminative with much less domain-specific information, and hence more invariant to domain shifts. 

Without duplicate elaboration, the target domain network holds the same architecture as the source domain network. In the second step of our approach, we fix the learned source domain factorization network and train the target factorization network with adversarial adaptation, as discussed in following section.

\subsection{Adversarial Domain Adaptation}

Our factorization network is designed to capture discriminant information in the task-specific subspace while dropping domain-specific information as much as possible. We leverage adversarial training to minimize the discrepancy between the task-specific subspace of the target domain and that of the source domain so that we can easily transfer the knowledge learned from source domain to target domain.  Specifically, we learn our target domain neural network by optimizing the following objective function:
\begin{equation}
\mathcal{L}_\text{target} = \mu \mathcal{L}_{\text{adv}_\mathbf{D}} + \nu \mathcal{L}_{\text{adv}_M}  + \mathcal{L}_{\text{r}}
\label{loss_total_target}
\end{equation}
where $\mu$ and $\nu$ are the hyper parameters that balance the contributions of adversarial training loss. 

The reconstruction loss $\mathcal{L}_\text{r}$ in target domain is similarly defined as Eqn.~\ref{recon_loss} over target domain network parameters. The adversarial training losses are defined similarly to the GAN loss~\cite{goodfellow2014generative}. Instead of using the task-specific subspace directly, we use the logit space obtained from the source domain to guide the learning in the target domain, which works better in practice. The discriminator $\mathbf{D}$ maps the input logit space into a binary label, where ``true" denotes the source domain and ``false" denotes the target domain. The target domain network is learned in an adversarial way to fool the discriminator so that the discrepancy between the two logit spaces is minimized. Specifically, the adversarial losses $\mathcal{L}_{\text{adv}_\mathbf{D}}$ for optimizing the discriminator $\mathbf{D}$ and $\mathcal{L}_{\text{adv}_M}$ for optimizing the target domain encoder are defined as

\begin{equation}
\begin{split}
\underset{\mathbf{D}}{\text{min}} \hspace{0.1cm} \mathcal{L}_{\text{adv}_\mathbf{D}} = - \mathbb{E}_{\mathbf{x}_s\sim \mathcal{S}} \hspace{0.1cm}\text{log}\mathbf{D}(M^s(\mathbf{h}_t^s; \theta_m^s))   -\mathbb{E}_{\mathbf{x}_t\sim \mathcal{T}} \hspace{0.1cm}\text{log}(1-\mathbf{D}(M^t(\mathbf{h}_t^t; \theta_m^t))
\end{split}
\end{equation}

\begin{equation}
\underset{\Theta}{\text{min}} \hspace{0.1cm} \mathcal{L}_{\text{adv}_M} =  - \mathbb{E}_{\mathbf{x}_t\sim \mathcal{T}} \hspace{0.1cm} \text{log}(\mathbf{D}(M^t(\mathbf{h}_t^t;\theta_m^t)))
\end{equation}
where $\Theta$ denote the network parameters for the target domain encoder and logit mapping. 
As the task-specific subspace at target domain aims to learn a similar distribution as the one from source domain, the mutual information loss is not necessary for the target domain. In the experiments, we did try using Eqn.~\ref{mutal_loss} at target domain , but did not observe further improvement.

Unlike the symmetric structure of our network as demonstrated in Figure~\ref{network}, we perform asymmetric adaptation during optimization where the target domain network is fine-tuned from source domain network instead of weight sharing for the two networks. Previous efforts explored using shared weights between source and target networks to reduce model parameters~\cite{ganin2016domain}\cite{tzeng2015simultaneous}, or leave the target network completely untied~\cite{rozantsev2016beyond}\cite{tzeng2017adversarial}. We found that it is not necessary to share the weights for shallow networks such as LeNet~\cite{lecun1998gradient}, but imperative to partially share some early network layers for deeper neural networks, such as ResNet~\cite{He2015}, which is the standard practice to train the deep nets. By jointly optimizing the adversarial loss and reconstruction loss, we force the target domain task-specific subspace to match the distribution of the source domain task-specific subspace, which is discriminative for the classification task, while leaving the less relevant target domain-specific representations for the domain-specific subspace to capture. Together, the two terms encourage the network to learn more discriminative and domain invariant feature representations for the task.

\section{Experiments}

We evaluate the proposed FAN on the tasks of unsupervised domain adaptation using benchmark datasets including MNIST~\cite{lecun1998gradient}, USPS~\cite{hull1994database} and SVHN~\cite{netzer2011reading}, as well as much larger real-world tagging datasets we collected that contain more than 100,000 images, respectively. We demonstrate that our approach is significantly improved compared to previous state-of-the-art methods.

\subsection{Digits Datasets}

\begin{figure}[!t]%
    \centering
    \subfloat[Example images from the three digits datasets. Left three columns: MNIST; middle three columns: USPS; right three columns: SVHN.\label{dataset_a}]{{\includegraphics[width=.46\linewidth]{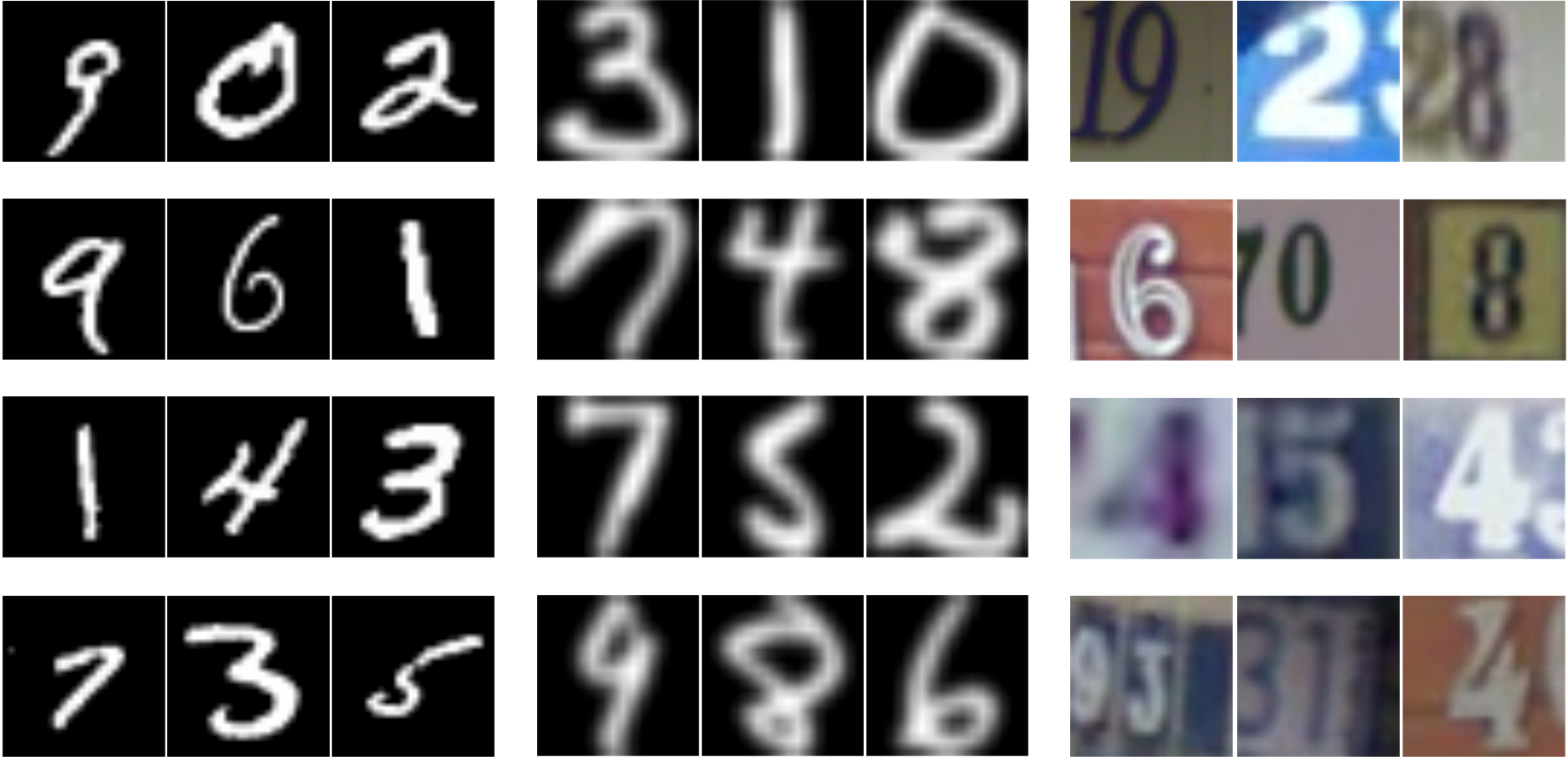} }}%
    \hfill
    \subfloat[Example images from the \textit{Crawling} dataset (left two columns) and the \textit{Mobile} dataset (right two columns). Top row: forest; bottom row: steering wheel.\label{dataset_b}]{{\includegraphics[width=.46\linewidth]{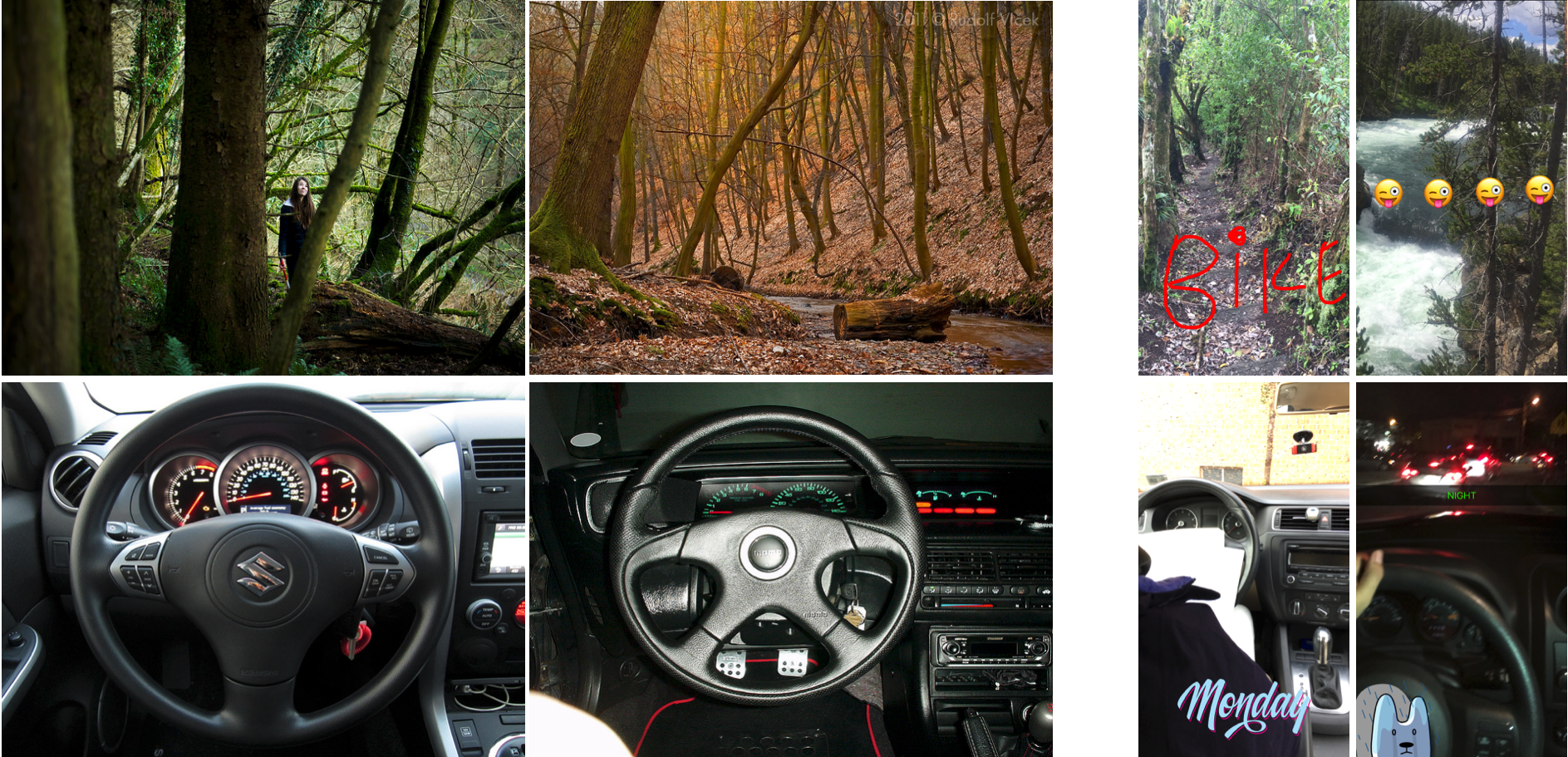} }}%
    \caption{Visualization of example images from the five datasets used in the study.}%
    \label{dataset}%
\vspace*{-3mm}
\end{figure}

We use three digits datasets, MNIST~\cite{lecun1998gradient}, USPS~\cite{hull1994database} and SVHN~\cite{netzer2011reading}, as the benchmark and follow the previous studies~\cite{ghifary2016deep}\cite{tzeng2015simultaneous}\cite{ganin2015unsupervised}\cite{tzeng2017adversarial}\cite{liu2016coupled} to perform three unsupervised adaptation settings including MNIST $\to$ USPS, USPS $\to$ MNIST and SVHN $\to$ MNIST. The benchmark datasets contain images of 10 digits ranging from 0 to 9. Some sample images from the three datasets are shown in Figure~\ref{dataset_a}. To run experiments in an unsupervised manner, the labels of the target domain training images are withheld. 

{\textbf{Network architecture}} 
The network we use in the experiments contains an encoder and a decoder and has the same structure under the three experiment settings. Following the recent work~\cite{tzeng2017adversarial} for fair comparison, we adopt a similarly modified LeNet~\cite{lecun1998gradient} as the
encoder that differs only in utilizing batch normalization (BN). We also applied BN for~\cite{tzeng2017adversarial} but observed no improvement. Specifically, the encoder consists of two convolutional layers with kernel size 5 and the number of filters 20 and 50, respectively. Each convolutional layer is followed by rectified linear units (ReLU), BN, and max pooling layers. After that we have two fully connected (FC) layers with 500 and 100 hidden units respectively. The activations from the last FC layer is split into two parts, one for domain-specific subspace and the other for task-specific subspace. The task-specific feature is connected to an FC layer to get the classification logits for prediction, while the domain-specific feature is concatenated with the classification logits as input for decoding phase. The decoder employs a deconvolution architecture~\cite{zeiler2010deconvolutional} including one FC layer with 300 hidden units, two 5$\times$5$\times$16 convolutional layers, one upsampling layer to 28$\times$28, and two 3$\times$3 convolutional layers with 16 and 1 filters, respectively. The FC layers and convolutional layers are followed by ReLU and BN, except for the last convolutional layer that gives the reconstruction output. The logit activations from the two domains are sent to the discriminator network which contains three FC layers. The first two FC layers have 500 hidden units followed by ReLU and BN. The last FC layer provides the domain label estimation for the input samples.  


{\textbf{Implementation details}}
Since images in different datasets varies in size, we resize the images in USPS and SVHN datasets to 28$\times$28 in order to match the input image size in MNIST. In addition, we convert the RGB images from SVHN to gray scale images. All the pixel values are normalized to a range of 0 to 1. For the unsupervised adaptation between MNIST and USPS, two training paradigms are implemented. The first one follows the training strategy introduced in~\cite{long2013transfer}, which sampled 2,000 training images from MNIST and 1,800 training images from USPS. For the second training protocol, we consider utilizing all the training data from the two domains and denote it as MNIST$\to$USPS (full) and USPS$\to$MNIST (full). For both training protocols, the testing set remains the same. For adaptation from SVHN to MNIST, we use all the training images from the two datasets. The training process contains two steps. The first step is to train a model in the source domain using Eqn.~\ref{loss_total} with $\alpha$ as 2 and $\beta$ as 1. In the second step, we fix the trained model in source domain 
and train the recognition model in the target domain using the Eqn.~\ref{loss_total_target}, where $\mu$ is 2 and $\nu$ is 1. We initialize the target domain network using the weights of the model trained in source domain. No data augmentation setting is utilized in the experiments.


\begin{table*}[!t]
\small
\caption{Experimental results on unsupervised domain adaptation for the digits  datasets including MNIST, USPS, and SVHN. Full denotes using the entire training set for the domain adaptation between MNIST and USPS. The last column shows the largest improvement over each method out of the three experiments.}
\vspace*{-3mm}
\begin{center}
\centerline{
\begin{tabular}{|  C{2.5cm} | C{2.3cm} | C{2.3cm} | C{2.35cm} |  C{2.2cm} |}
\hline
{Method} & MNIST$\to$USPS & USPS$\to$MNIST & SVHN$\to$MNIST & Largest Improvement\\ \hline\hline
{Baseline} &  0.752 $\pm$ 0.016 & 0.571 $\pm$ 0.017 & 0.601 $\pm$ 0.011 & 0.339\\ \hline
{DSN\cite{bousmalis2016domain}\cite{bousmalis2017unsupervised}} & 0.913 & - & 0.827 & 0.098\\ \hline
{RevGrad\cite{ganin2015unsupervised}} &  0.771 $\pm$ 0.018 & 0.730 $\pm$ 0.020 & 0.739 &0.186 \\ \hline
{DDC\cite{tzeng2015simultaneous}} & 0.791 $\pm$ 0.005 & 0.665 $\pm$ 0.033 & 0.681 $\pm$ 0.003 &0.245 \\ \hline
{CoGAN\cite{liu2016coupled}} & 0.912 $\pm$ 0.008 & 0.891 $\pm$ 0.008 & - &0.019 \\ \hline
{DRCN\cite{ghifary2016deep}} & 0.918 $\pm$ 0.0009 & 0.737 $\pm$ 0.0004 & 0.820 $\pm$ 0.0016&0.173 \\ \hline
{ADDA\cite{tzeng2017adversarial}} & 0.894 $\pm$ 0.002 & 0.901 $\pm$ 0.008 & 0.760 $\pm$ 0.018&0.165\\ \hline
{Ours} & \textbf{0.921 $\pm$ 0.014} & \textbf{0.910 $\pm$ 0.011} & \textbf{0.925 $\pm$ 0.011}&-\\ \hline
{Ours (full)} & \textbf{0.963 $\pm$ 0.002} & \textbf{0.971 $\pm$ 0.008} & -&-\\ \hline
\end{tabular}
}
\end{center}
\label{soa}
\end{table*}
{\textbf{Comparison results}}
Table~\ref{soa} shows our results as compared with recent methods. Our approach clearly achieves the best overall performance on all three domain adaptation experiments under the same settings. Compared with previous methods, our method significantly outperforms each of them at least on one of the three experiments, with a gap of over 10\% in many cases, as shown in the last column in Table~\ref{soa}. For the adaptation between MNIST and USPS, we also show results using the full set of training data from both domains and observe that it significantly improves the accuracy, implying that our adaptation network can better minimize the distribution shift with more training data. 

\begin{figure}[!t]
\begin{center}
\includegraphics[width=0.7\columnwidth]{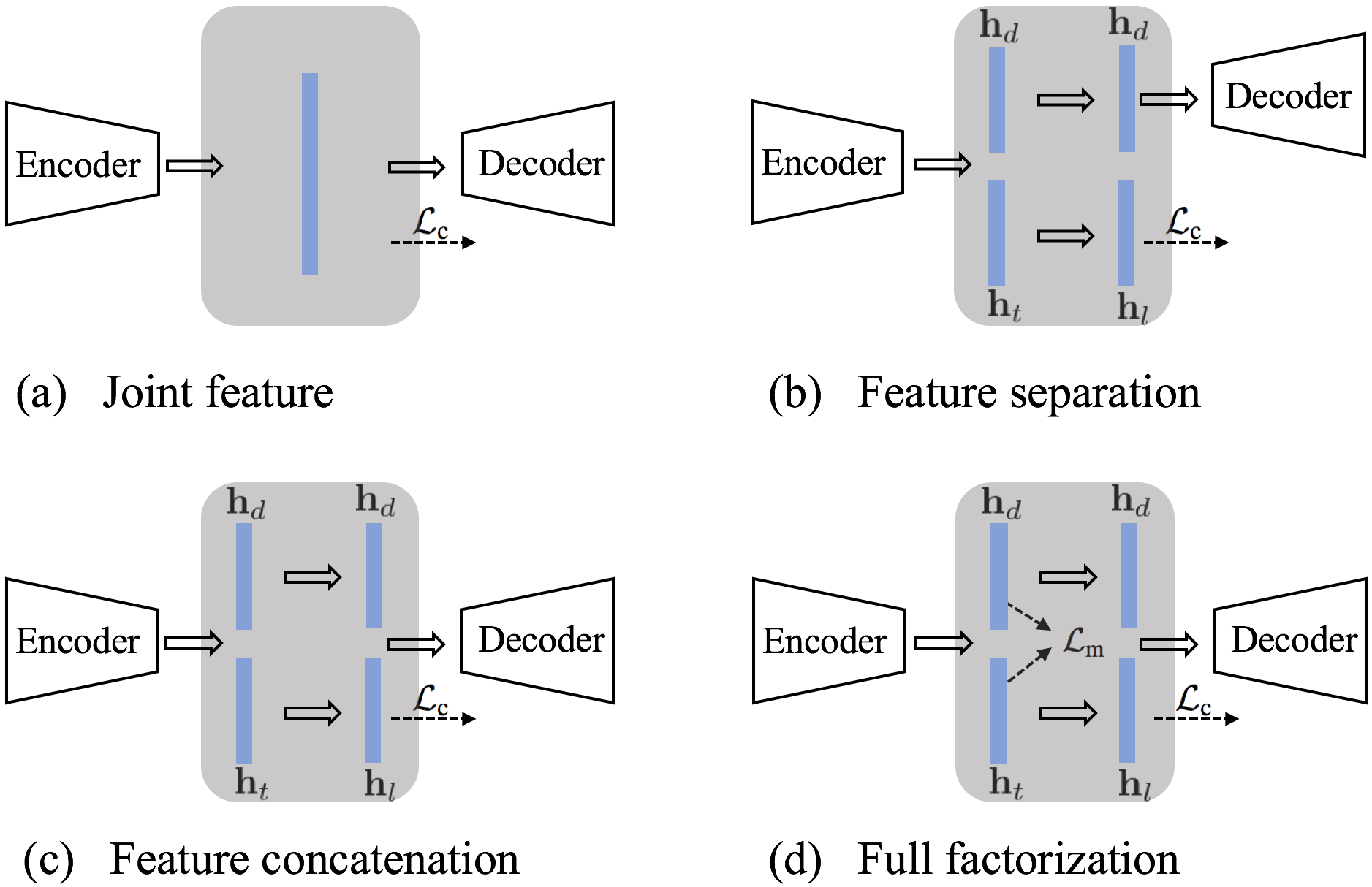}
\caption{Four network architectures for the study of feature factorization.}
\label{four_figs}
\end{center}
\vspace*{-3mm}
\end{figure}

{\textbf{Ablation analysis of our network design}}
We conduct ablation study on the design of our factorization architecture. The structure for four network settings are shown in Figure~\ref{four_figs} with the following details. 
\begin{itemize}
\item \textbf{Joint feature}: As shown in Figure~\ref{four_figs}a, we learn a joint feature space for both image reconstruction and classification, and use reconstruction losses in both domains along with the classification loss in source domain to train the network.
\item \textbf{Feature separation}: As shown in Figure~\ref{four_figs}b, in this setting, we separate the latent features into two parts. One part is used for reconstruction and the other part is used for classification. 
\item \textbf{Feature concatenation}: As shown in Figure~\ref{four_figs}c, the previous reconstruction features are concatenated with the classification logits as new reconstruction features. 
\item \textbf{Full factorization}: As shown in Figure~\ref{four_figs}d, we add mutual information loss in this setting to explicitly enforce the orthogonality between the two separated features, thus factorizing the latent feature space into a domain-specific subspace and a task-specific subspace.
\end{itemize}

For all four settings, we conduct the same two-stage training process and apply the adversarial learning at the second stage.
The results shown in Table~\ref{ablation} indicate that we could obtain stronger results by better separating the features, and our factorization method yields the best results. 

\begin{figure}[!t]%
    \centering
    \subfloat[Embedding of the logit space before the adaptation.\label{quantitative_a}]{{\includegraphics[width=.22\linewidth]{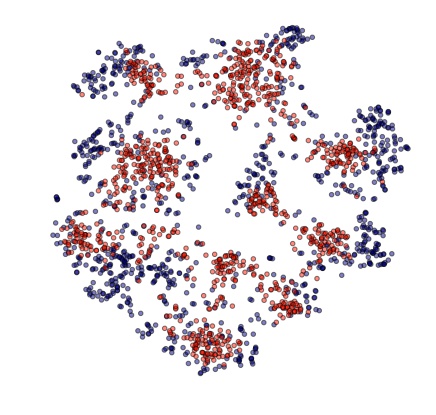} }}%
    \hfill
    \subfloat[Embedding of the logit space after the adaptation.\label{quantitative_b}]{{\includegraphics[width=.22\linewidth]{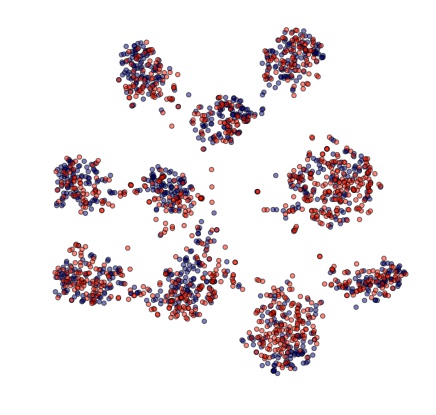} }}%
    \hfill
    \subfloat[Embedding of the domain-specific subspace before the adaptation.\label{quantitative_c}]{{\includegraphics[width=.22\linewidth]{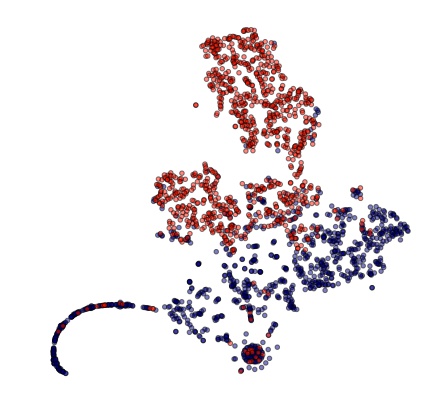} }}%
    \hfill
    \subfloat[Embedding of the domain-specific subspace after the adaptation.\label{quantitative_d}]{{\includegraphics[width=.22\linewidth]{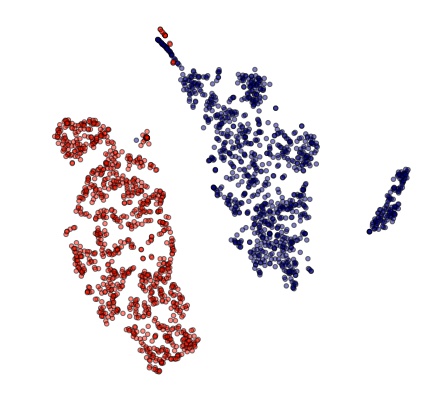} }}%
    \caption{Visualization of the domain adaptation from SVHN (source domain, red color) to MNIST (target domain, blue color). We show the visualization of t-SNE embedding for the logits space before adaptation (a) and after adaptation (b), and the domain-specific subspace before adaptation (c) and after adaptation (d). }%
    \label{quantitative}%
\end{figure}

\begin{table*}[!b]
\small
\caption{Analysis of the effects of feature factorization under different network structures.}
\begin{center}
\centerline{
\begin{tabular}{|  C{3.4cm} | C{2.2cm} | C{2.2cm} |C{2.2cm} |C{2.2cm} |}
\hline
Method & Joint feature  & Feature separation & Feature concatenation & Full factorization\\ \hline\hline
MNIST $\to$ USPS (full) &  0.955 $\pm$ 0.004 & 0.958 $\pm$ 0.002 & 0.961 $\pm$ 0.002 & \textbf{0.963 $\pm$ 0.002}\\ \hline
USPS $\to$ MNIST (full) & 0.933 $\pm$ 0.017 & 0.936 $\pm$ 0.014 & 0.958 $\pm$ 0.009 & \textbf{0.971 $\pm$ 0.008}\\ \hline
SVHN $\to$ MNIST &  0.829 $\pm$ 0.019 & 0.858 $\pm$ 0.024 & 0.905 $\pm$ 0.006 & \textbf{0.925 $\pm$ 0.011}\\ \hline
\end{tabular}
}
\end{center}
\label{ablation}
\end{table*}
{\textbf{Analysis of the embedding spaces}}
Besides the quantitative results, we visualize the high-dimensional features of the factorized subspaces in the 2D plane for adaptation from SVHN to MNIST using the t-SNE~\cite{maaten2008visualizing}. 
We randomly select 1,000 images from the two testing sets and show visualization results in Figure~\ref{quantitative}. We set perplexity to 35 for all four visualization results. 
The embedding of the logits space before and after adaptation for the two domains are shown in Figure~\ref{quantitative_a} and Figure~\ref{quantitative_b}, respectively. As expected, after adaptation, the samples from the target domain are clustered into more obvious groups and match better with the clusters in the source domain.

The visualization of the domain-specific subspaces before and after adaptation are shown in Figure~\ref{quantitative_c} and Figure~\ref{quantitative_d}, respectively. 
After adaptation, we simultaneously learn a good task-specific subspace on the target domain and a good domain-specific subspace. The domain-specific subspace should capture information specific to the domain, and therefore, the two domain-specific subspaces are further divided after adaptation, which proves our learning algorithm is effective.

\begin{figure}[!b]%
    \centering
    \subfloat[\label{recon_cat_a}]{{\includegraphics[width=.4\linewidth]{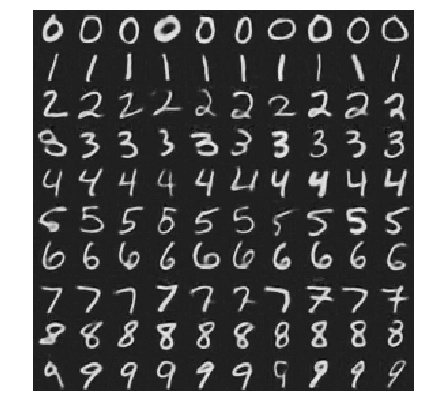} }}%
    \subfloat[\label{recon_cat_b}]{{\includegraphics[width=.4\linewidth]{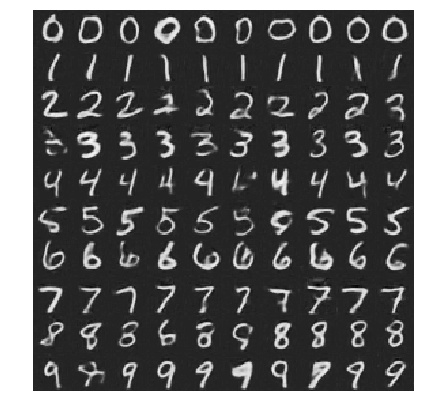} }}%
    \vspace*{-3mm}
    \caption{Reconstruction results using the target domain reconstruction network for domain adaptation from SVHN to MNIST. (a) Reconstruction results using the testing samples from target domain. (b) Reconstruction results using the concatenation of domain specific features from target domain and classification logits from source domain.}%
    \label{recon_cat}%
\end{figure}

Furthermore, we analyze the embedding subspaces of the target domain by showing two reconstruction results. Figure~\ref{recon_cat_a} demonstrates the reconstruction results using features extracted from target domain testing samples. We also concatenate the domain-specific features of the target domain samples with the logits activations of randomly selected testing images from source domain of the same class, and show the reconstruction results in Figure~\ref{recon_cat_b}.
Although reconstruction quality of Figure~\ref{recon_cat_b} is not as good as that of Figure~\ref{recon_cat_a}, the images are still very similar, which proves that task-specific subspaces for the two domains indeed share similar distributions and that the target domain-specific subspace stores the domain characteristics for reconstruction.


\subsection{Real-world tagging Datasets}

While many studies in the literature tackle unsupervised domain adaptation, they mostly evaluate their algorithms on small and simple datasets such as the digits dataset~\cite{lecun1998gradient}\cite{hull1994database}\cite{netzer2011reading}, and the office dataset~\cite{griffin2007caltech}. The capacity for domain adaptation algorithms to work for large-scale real-world complex applications remains unclear. Previous work \cite{bousmalis2016domain} points out some problems with evaluation on office dataset~\cite{griffin2007caltech}\cite{saenko2010adapting}, where pretrained models from ImageNet have to be used~\cite{sun2016return}. So instead of working on the toy office dataset, we collected two real-world tagging datasets to benchmark unsupervised domain adaptation algorithms, where we have sufficient images to train deep networks from scratch.

The first dataset is  collected from the search engines and named \textit{Crawling} dataset, while the second dataset is collected from the photos shot by mobile phones and titled \textit{Mobile} dataset. The two datasets contain the same 100 classes. Some example images from the two datasets are shown in Figure~\ref{dataset_b}.
There are two major differences between the two datasets: 1) the images in \textit{Crawling} dataset usually have good quality and clear background while the images in the \textit{Mobile} dataset suffer from several defects such as image blur and out of focus, as well as noisy background and various image filters and stickers; 2) the \textit{Mobile} dataset contains mostly vertical images while the images from the \textit{Crawling} dataset have various image ratios.  
We use the \textit{Crawling} data as the source domain and the \textit{Mobile} data as the target domain because we can easily collect crawling data with labels by keyword searching.
The \textit{Crawling} dataset includes 150,000 training images. The \textit{Mobile} dataset contains 115,000 images out of which we randomly select 100,000 images as the training set, 10,000 images as the testing set, and others as the validation set. 
Compared with the digits datasets, the real-world tagging datasets not only have a larger scale but also are more suitable for the study on real-world scenarios.

{\textbf{Network architecture}}
The encoder part of our network uses the ResNet-50~\cite{He2015} architecture. The activations from the last average pooling layer are factorized equally into two parts. The task-specific subspace features are followed by a FC layer to estimate the classification logits, and the domain-specific features are concatenated with the classification logits to serve as the input for the decoder.
The decoder network uses architecture from DCGAN~\cite{radford2015unsupervised}. It contains 5 fractionally-strided convolutions layers with 256, 256, 128, 64 and 3 filters respectively. Each layer is followed by ReLU and BN, except for the last layer.
The discriminator network contains three FC layers. The first two FC layers have 1024  and 2048 hidden units respectively, followed by ReLU and BN. The last FC layer output is used for label domain classification.  

{\textbf{Implementation Detail}}
All images are resized to 256$\times$256 and randomly cropped to 224$\times$224 during the training process. 
$\alpha$ is set to 5 and $\beta$ is set to 1 for  Equation~\ref{loss_total}. And we set $\mu$ as 2 and $\nu$ as 1 for  Eqn.~\ref{loss_total_target}. 

In order to measure whether more unlabeled training images in target domain would contribute to the generalizability of the target model, we perform three sets of experiments in addition to that without adaptation. In the first two sets, we randomly select 10\%  and 50\% images from each class of the target training set, while in the third set, we use the full target training set. The Top-1 and Top-5 accuracy for the testing set of the target domain are shown in Table~\ref{snap_result}. Compared with the model without adaptation, using 10\% training images from each class could improve the the Top-1 and Top-5 accuracy as 3.75\% and 3.69\% respectively. Using the full training set improves the Top-1 accuracy by more than 10\% and Top-5 accuracy more than 12\%.
We also compared our results with ADDA~\cite{tzeng2017adversarial} using the full target training set, as shown in Table~\ref{snap_result}. Our approach outperforms ADDA on both Top-1 and Top-5 accuracy as 2.46\% and 3.05\% respectively.
These results demonstrate that our method can significantly improve the performance over baselines in real-world applications. In addition, we show that more unlabeled training data from the target domain helps the unsupervised adaptation.
\vspace*{-3mm}
\begin{table*}
\small
\caption{Top-1 and Top-5 accuracies on the testing set of the \textit{Mobile} dataset.}
\label{snap_result}
\begin{center}
\centerline{
\begin{tabular}{|  C{5.5cm} | C{1.2cm} | C{1.2cm} |}
\hline
Method & Top-1  & Top-5  \\ \hline\hline
No adaptation & 0.3571 & 0.6607 \\ \hline
ADDA\cite{tzeng2017adversarial} (full set of target training)& 0.4386 & 0.7533 \\ \hline\hline
Ours (10\% of target training) & 0.3946 & 0.6976 \\ \hline
Ours (50\% of target training)& 0.4041 & 0.7018 \\ \hline
Ours (full set of target training)& \textbf{0.4632} & \textbf{0.7838} \\ \hline
\end{tabular}
}
\end{center}
\vspace*{-8mm}
\end{table*}

\section{Conclusion}

In this paper, we introduce FAN for unsupervised domain adaptation. We factorize the latent feature space into task-specific subspace and domain-specific subspace for both source and target domains and consider the domain adaptation only on task-specific subspace. The network in source domain is jointly trained with image classification and reconstruction under the factorization architecture to learn the discriminative task-specific subspace while pushing away domain-specific information as much as possible. The network in target domain is learned under the same factorization structure with GAN loss to adapt the target domain task-specific subspace to the source domain task-specific subspace. We evaluate our proposed framework on four domain adaptation tasks, all achieving state-of-the-art results. For future work, we would like to extend our algorithm to other vision tasks beyond image classification.


\clearpage

\bibliographystyle{splncs}
\bibliography{ms.bbl}
\end{document}